%% file: ShapeDNA.tex
\documentclass[10pt,journal,cspaper,compsoc]{IEEEtran}
\ifCLASSOPTIONcompsoc
\else
\fi

\ifCLASSINFOpdf
   \usepackage[pdftex]{graphicx}
\else
\fi

\usepackage{wrapfig}

\usepackage{cite}
\usepackage{url}

\usepackage{amsmath}

\usepackage{color} 
\usepackage{multirow,times}

\usepackage{epstopdf} 

\mathchardef\mhyphen="2D

\begin{document}

\bstctlcite{IEEEexample:BSTcontrol}
%
\title{Shape DNA: Basic Generating Functions \\ for Geometric Moment Invariants}
%
%
%
%

\author{Erbo Li,
        Yazhou Huang,
        Dong Xu and Hua Li
\IEEEcompsocitemizethanks{\IEEEcompsocthanksitem H. Li is with Key lab of Intelligent Information Processing, Institute of Computing Technology, Chinese Academy of Sciences. 
E-mail: lihua@ict.ac.cn
}
}

%
%


\markboth{}%
{Shell \MakeLowercase{\textit{Erbo Li et al.}}: Shape DNA: Basic Generating Functions for Geometric Moment Invariants}

\IEEEcompsoctitleabstractindextext{%
\begin{abstract}

Geometric moment invariants (GMIs) have been widely used as basic tool in shape analysis and information retrieval. Their structure and characteristics determine efficiency and effectiveness. Two fundamental building blocks or generating functions (GFs) for invariants are discovered, which are dot product and vector product of point vectors in Euclidean space. The primitive invariants (PIs) can be derived by carefully selecting different products of GFs  and calculating the corresponding multiple integrals, which translates polynomials of coordinates of point vectors into geometric moments. Then the invariants themselves are expressed in the form of product of moments. This procedure is just like DNA encoding proteins. All GMIs available in the literature can be decomposed into linear combinations of PIs. This paper shows that Hu's seven well known GMIs in computer vision have a more deep structure, which can be further divided into combination of simpler PIs. In practical uses, low order independent GMIs are of particular interest. In this paper, a set of PIs for similarity transformation and affine transformation in 2D are presented, which are simpler to use, and some of which are newly reported. The discovery of the two generating functions provides a new perspective of better understanding shapes in 2D and 3D Euclidean spaces, and the method proposed  can be further extended to higher dimensional spaces and different manifolds, such as curves, surfaces and so on.


\end{abstract}

\begin{IEEEkeywords}
geometric moment invariant, generating function, primitive invariant,shape DNA, affine invariant, independent invariant, 3D moment invariant.
\end{IEEEkeywords}}

\maketitle

\IEEEdisplaynotcompsoctitleabstractindextext

%
\IEEEpeerreviewmaketitle

\section{Introduction}
\label{sec:intro}

\input{intro}


\input{related_work}

\input{GMI}

\input{Extension}
\input{independent}

\input{Conclusion}


\appendix 
\input{Appendix}

\section*{Acknowledgment}

This work was partly funded by National Natural Science Foundation of China (Grant No. 60873164, 61227082 and  61379082).

\ifCLASSOPTIONcaptionsoff
  \newpage
\fi


\bibliographystyle{IEEEtran}
\bibliography{ShapeDNA}
\end{document}

%% file: intro.tex

\IEEEPARstart{I}{t} has been proven difficult to find intrinsic characteristics of shapes that remain unchanged after geometric transformations, especially when dealing with real world problems.
In recent years, geometric moment invariants (GMIs) in the form of polynomials of different order of moments have been well studied in the research of shape analysis, however this may not work well in practical use.
One of the reasons is that the number of useful invariants is quite limited, they could not fully describe the geometric characteristics of shapes.
Another reason is that when the degree of invariants as a polynomial is high, the calculation of invariants is not stable enough to get desirable calculation results.

Theoretically the structure of invariants is an interesting problem. And practically it is desirable to find enough low degree or lower order invariants as useful shape descriptors.
In this paper, we focus on exploring the structure of invariants and finding such a set of invariant descriptors for shape analysis.

First, we propose two fundamental generating functions (GFs) as the ``DNA'' of shapes. They are in the form of dot product and determinant composed of point vectors. In 2D space the determinant is equivalent to the cross product.
In theory, an infinite number of invariant definitions could be derived as the product of a series of GFs.
By carefully selecting GFs and using multiple integrals, corresponding GMIs can be explicitly constructed, including known GMIs in the existing literature and unknown ones.
With this method, we have proposed a group of independent GMIs as shape descriptors.

Second, by using GFs analysis, we have found that Hu's seven GMIs can be further decomposed into a simpler set of Primitive Invariants (PIs), leading to a new level of invariant study.

Third, our method can be applied to both 2D and 3D spaces, and further extended to higher dimensional spaces. It is also applicable to different manifolds, including curves and surfaces.

To summarize, we have found the GFs that can encode GMIs and further construct shape descriptors, just like DNA encodes proteins.
Also the properties of shapes are determined by the selection of GFs, similar to protein's properties being determined by the structure of DNA. With these GFs, we can re-express and perform in-depth analysis on the known GMIs from existing literature, and also search for new GMIs. Therefore in a way we claim to have found the Shape DNA.

%% file: related_work.tex

\section{Related Work}
It has been an interesting research topic to unveil the essence of shapes. For years many attempts had been made to find the basic units that determine the unchanged characteristics of shapes, i.e. invariants, when transformations of a certain type are applied to the shapes. A direct application is to recognize and compare shapes that have gone through different transformations. The real world study subjects include 2D images and 3D objects.

Extracting shape characteristics independent of transformations is not an easy task. The use of geometric moment invariants could be a good solution. Our goal is to find such a group of invariants, in the simplest form, independent with each other within the group, that can express all possible invariants. Therefore such a group of invariants can be used as shape descriptors. It is able to solve a wide range of problems related to shapes including image processing, shape analysis, object recognition, scene analysis, pattern classification, shape retrieval, etc.

The study of invariants is built on solid mathematical background. Broadly speaking, Hilbert \cite{Hilbert1893} had proved that invariants only have limited bases, and the rest can be expressed by the bases. His theory provides the basis for the topic of invariants, and indicates a potential way of analyzing structure of invariants. Although his invariants is mathematically different from GMIs, with that knowledge it is possible for us to find GMIs structure or all the bases, which could be the basic units of shape descriptors.

The acquisition of GMIs is a quite difficult task. Hu \cite{Hu1962} first discussed the invariants defined on images, and proposed seven well-known moment invariants under general two-dimensional linear transformations or affine transformations. Since then, a large number of works have been devoted to improve or broaden the concept of invariants in many different ways. Teh and Chin\cite{Teh1988} defined various types of moments for shape description and image compression. Li \cite{Li1992} found fifty-two 2D moment invariants up to order nine. Suk and Flusser \cite{Flusser2000} applied the concept of independence to geometric invariants, and proposed a complete set of independent invariants for rotation transformations. Structure moment invariants proposed by Li et al \cite{ZML2005} increased the difference of similar objects by changing the density function.

The study of GMIs was soon extended from 2D to 3D and higher dimensions. Sadjadi and Hall \cite{Sadjadi1980} extended the concept of invarinats to 3D for shape analysis, and provided geometrical interpretation of moment invariants. Lo and Don \cite{LoD89} constructed invariants by complex moments and group theory. Tensors theory \cite{Reiss92} could also be used for generation of moment invariants in 2D and 3D space. 3D polar-radius invariant moment was proposed for the 3D object recognition and classification \cite{3DZongminLi2005}. A fast depth-buffer-based voxelization algorithm had been proposed in \cite{Karabassi1999}. 

In addition, the invariants have been used in a wider range of applications.
Brain morphometry using 3D GMIs was proposed in \cite{Mangin2004}. Characterizing fMRI activations within regions of interest using 3D GMIs was introduced in \cite{Ng2006}. Deriving 3D GMIs and using them as shape descriptors for the representation of 3D models were described in \cite{Zongkai2006}.

The goal following that trend to increase definition dimensions was to find a general method to generate GMIs, since some methods were applicable for 2D moment invariants but failed to extend to higher dimensions. Mamistvalov \cite{Mamistvalov98} selected several n-dimensional moment invariants for pattern recognition. Flusser et al \cite{Flusser2003} constructed rotational invariant by group representation theory similar to \cite{LoD89}.

The definition of GMIs had also be applied to different transformations. It started with similarity transformations \cite{Hu1962}\cite{Li1992}, including rotation, translation and uniform scaling. Later the transformations had been extended to affine transformations by Suk and Flusser \cite{Flusser1993}, and generating graphs had been used to express 2-D affine GMIs \cite{Flusser2004},\cite{Flusser2011}. Xu and Li \cite{XuL2006} proposed a geometric method for constructing GMIs, providing a solid basis for this paper. Since affine transformation is a special case of projective transformation, it is generally used to approximate the projective transformation and avoid dealing with the complicated non-linear case. The projective transformation is the most common case in real world, many attempts have been made over the years to find projective moment invariants.  Suk and Flusser \cite{Suk04} tried to define projective moment invariants in the form of infinite series, which may not be a convincing solution as determined affine moment invariants. Other attempts were made to define projective invariants under strict constraints \cite{Yuanbin2008}, \cite{Yuanbin2009}, \cite{Yuanbin2010}, and Wang et al. gave a sound solution in \cite{Wang2015} using partial derivatives. 

Another direction was to define GMIs on different manifolds or domains. Since moment invariants are considered a type of integral invariants, it is possible to set the integral domain to be one part of the entire object to solve partial/local shape description and recognition problems, like a segment of a curve or a part of a surface. In 2D space, curve moments had been proposed to solve partially occluded object recognition \cite{Sluzek1988}. And in 3D space, there are three kinds of manifolds: solid, curve and surface. Xu and Li  computed moments based on 3D solid \cite{Xu2008}, defined 3D curve moment invariants for the representation of parametric curves \cite{Li2006}, and proposed 3D surface moment invariants for the representation of free-form surfaces \cite{Xu2006}, which can be used in many applications like remodeling and reconstruction of human face.

In recent years, the study of invariants has been extended to broader areas, and the concept of invariants has been further developed. Gong defined a type of color descriptor by color affine moment invariants \cite{Gong2013}, similar as the shape descriptors in Euclidean space. Flusser \cite{Flusser2009} presented a survey of image analysis and pattern recognition methods based on image moments, demonstrating moment invariants in real applications across computer vision, remote sensing and medical imaging. More recently Papakostas \cite{Papakostas2014} collected new developments in this area of research. Karakasis \cite{Karakasis2015} had found eight independent 2D affine invariants, however not with lowest order. We will address in this paper the definition of independent invariants in the lowest order.

Our constructing method is inspired by Xu and Li \cite{Xu2008}, who constructed moment invariants by using multiple integrals of invariant primitives like distance, angle, area and volume. It provided an intuitive way to define invariants using geometric entities, and good extensibility.
In this paper, we further simplify the set of geometric entities, and propose only two generating functions for the construction of invariant cores. By carefully selecting the combination of generating functions as the invariant cores, and applying them to multiple integrals, we are able to directly compose different moment invariants. By re-express 2D moment invariants in prior arts, we have validated our generating method. Also, our method is applicable to moment invariants under different transformations, and we have proposed a complete and simplest set of affine moment invariants in specified degree and order of moments. Our method is easy to extend to higher dimension and manifolds. We provide the definition of generating functions in 3D, and can re-express all the existing 3D moment invariants.  

Since moment invariants can be used to describe characteristics of shapes, we have found the basic units of shape descriptors. Also the construction of moment invariants requires multiple integrals, it experiences a translation process, similar as DNA encode proteins. Therefore, we have found the ``DNA of shapes''.

%% file: GMI.tex

\section{Geometric Moment Invariants}
\label{sec:GeometricMomentInvariants}

\subsection{Definition of moments and invariants}

For convenience to discussion, we give some basic definitions here.

\textbf{Definition 1. Moment}

Let $m_{pq}$ denote the geometric moment in 2D dimension, i.e. on a 2D image, where $p$ and $q$ are non-negative integers, see Eq.~\eqref{eq:moment}.
The sum of $p$ and $q$ represents the order of the moment, and $l(x,y)$ represents image intensity at point (i.e. pixel) $(x,y)$.

\begin{equation}
\label{eq:moment}
	m_{pq} = \int^{\infty}_{-\infty}\int^{\infty}_{-\infty} x^{p}y^{q}l(x,y) dxdy
\end{equation}

Similarly, $m_{pqr}$ denotes the 3D moment with an order of $(p+q+r)$ , shown in Eq.~\eqref{eq:3D}.

\begin{equation}
\label{eq:3D}
	m_{pqr} = \int^{\infty}_{-\infty}\int^{\infty}_{-\infty}\int^{\infty}_{-\infty} x^{p}y^{q}z^{r}l(x,y,z) dxdydz
\end{equation}

\textbf{Definition 2. Central moment}

The centroid of 2D image can be determined by the zeroth and the first order moments, respectively:

\begin{equation}
\label{eq:centroid}
	\bar{x} = \frac{m_{10}}{m_{00}}, \  \bar{y} = \frac{m_{01}}{m_{00}}
\end{equation}

By translating the coordinates to the centroid, we define the central moment $\mu_{pq}$ as Eq.~\eqref{eq:central-moment}.
Since such translation can be achieved fairly easily, in the paper we mainly focus on the central moments.

\begin{equation}
\label{eq:central-moment}
	\mu_{pq} = \int^{\infty}_{-\infty}\int^{\infty}_{-\infty} (x - \bar{x})^{p}(y - \bar{y})^{q}l(x,y) dxdy
\end{equation}

\textbf{Definition 3. Invariant}

Given a function $I(a_1, a_2, ... , a_n)$ with parameters $a_j \in R^1$, suppose under some geometric transformation $T$, the transformed function  $I^{'}(a_1^{'}, a_2^{'}, ... , a_n^{'})$, with new parameters $a_j^{'}$.
If Eq.~\eqref{eq:GMI} holds,

\begin{equation}
\label{eq:GMI}
	I^{'}(a_1^{'}, a_2^{'}, ... , a_n^{'})= w^{k}I(a_1, a_2, ... , a_n) 
\end{equation}

where $w$ related to transformation $T$ and $k$ is an integer, then function $I(a_1, a_2, ... , a_n)$ is called invariant.
Specifically when $k = 0$, $I$ is absolute invariant; otherwise it is a relative invariant.
Note that the invariant $I$ is always related to a transformation, because according to the Erlangen Program \cite{klein2004elementary}, different transformations define or induce different invariants. 
In this paper, our focus is on the most common encountered geometric transformations, for example similarity transformations (including rotation, scale and translation) and affine transformation.

\subsection{Generating functions}
Xu and Li in \cite{Xu2008} presented an intuitive way of deriving moment invariants, and found three geometric primitives shown in Eq.~\eqref{eq:2-D primitive}, where 
$D(O,i)$, $R(O,i,j)$ and $A(O,i,j)$ represent the distance between two points, the angle between two vectors, and the triangular area formed by three vertices, respectively.
Since the focus of this paper is on central moments, the origin $O$ is used as one of the points.

\begin{equation}
\begin{aligned}
\label{eq:2-D primitive}
D(O,i) &=  (x^{2}_{i} + y^{2}_{i})^{\frac{1}{2}} \\
R(O,i,j) &= (x_{i},y_{i}) \cdot (x_{j},y_{j}) \\
A(O,i,j) &=  \frac{1}{2} (\left\|(x_{i},y_{i})\times(x_{j},y_{j}) \right\|)^{2}
\end{aligned}
\end{equation}

Through multiple integrals of invariant geometric primitives, a large portion of the moment invariants available in the literature can be derived directly, and also more new invariants can be obtained without the need to consider more complicated mathematical tools.

By analyzing the geometric primitives above, we have found a way to further simplify them. For calculability and convenience, the distance function $D$ always appears in the form of the square, and the area function $A$ can be expressed as order two determinant, the constant $1/2$ is omitted.

Here we define two generating functions $f(i,j)$ and $g(i,j)$, shown in Eq.~\eqref{eq:dot-product} and Eq.~\eqref{eq:cross-product}.
They represent dot-product and cross-product of two vector points in a given image $l(x,y)$ respectively.

\begin{equation}
\label{eq:dot-product}
	f(i,j) = (x_{i},y_{i})\cdot(x_{j},y_{j}) = x_{i}x_{j} + y_{i}y_{j}
\end{equation}

\begin{equation}
\label{eq:cross-product}
g(i,j) = 
\begin{vmatrix}
x_{i} \ y_{i} \\
x_{j} \ y_{j}
\end{vmatrix}
= x_{i}y_{j} - x_{j}y_{i}
\end{equation}

With the definition of two generating functions, $D$, $R$ and $A$ can be written as
$f(i,i)$, $f(i,j)$ and $g(i,j)$.

It is not a coincidence that we have chosen dot-product and cross-product as generating functions to generate invariants. Dot-product is also called inner-product, describing the isometric characteristics of an object itself.
While cross-product, as the outer characteristics, describes the selection of coordinates. Together they covered all the necessary and intrinsic information for shape analysis.

And $g$ function is a second order determinant, which is also the area without coefficient $1/2$.
In this way, the two functions can also be expressed as inner product of points vectors and the determinant defined by the point vectors, which is more convenient to extend to higher dimensional spaces. 
The two functions act as the fundamental building blocks for the definition of geometric invariants and for the description of shape spaces, which provides a deep insight into the inner structure of different shapes.

\subsection{Translation: from generating functions to invariants}
\label{sec:Translation}

Here we use an example to explain how to translate generating functions into GMIs.

In \cite{Hu1962}, Hu first introduced his seven famous invariants. The first invariant $I_1$ is

\begin{equation}
\label{eq:Hu1}
I_1 = \mu_{20} + \mu_{02}
\end{equation}

It can be re-expressed with our generating functions.
Applying Eq.~\eqref{eq:dot-product} by using a single point twice, we could get Eq.~\eqref{eq:f1_1}.

\begin{equation}
\label{eq:f1_1}
f(1,1) = x_{1}x_{1} + y_{1}y_{1}
\end{equation}

By definition in Eq.~\eqref{eq:central-moment} where the centroid is the origin $O$, then the central moments are shown in Eq.~\eqref{eq:Hu1_translation}, which is equivalent to Eq.~\eqref{eq:Hu1}.

\begin{equation}
\begin{aligned}
\label{eq:Hu1_translation}
&\ \ \ \ \ \int^{\infty}_{-\infty}\int^{\infty}_{-\infty} f(1,1)l(x,y) dxdy \\
&= \int^{\infty}_{-\infty}\int^{\infty}_{-\infty} (x_{1}x_{1} + y_{1}y_{1})l(x_1,y_1) dx_{1}dy_{1} \\
&= \int^{\infty}_{-\infty}\int^{\infty}_{-\infty} x_{1}^{2}l(x_1,y_1) dx_1dy_1 + \\
& \ \ \ \int^{\infty}_{-\infty}\int^{\infty}_{-\infty} y_{1}^{2}l(x_1,y_1) dx_1dy_1 \\
&= \mu_{20} + \mu_{02} \\
&= I_1 \\
\end{aligned}
\end{equation}

We use the symbol $\Leftrightarrow$ to denote the mapping between the invariants $I$ and the combination of generating functions.
For any given GMI, its corresponding generating functions can be found. Conversely, by carefully selecting the combination of generating functions, we can derive the corresponding GMI.

\begin{equation}
\label{eq:arrow}
I_1 \Leftrightarrow f(1,1)
\end{equation}

Following our example, Hu's seven invariants \cite{Hu1962} listed in Eq.~\eqref{eq:Hu} could then be re-expressed using solely the generating functions, shown in Eq.~\eqref{eq:Hu-generate}.

\begin{equation}
\begin{aligned}
\label{eq:Hu}
I_1 &= \mu_{20} + \mu_{02} \\
I_2 &= (\mu_{20} - \mu_{02})^2 + 4 \mu_{11}^2 \\
I_3 &= (\mu_{30} - 3\mu_{12})^2 + (3\mu_{21} - \mu_{03})^2 \\
I_4 &= (\mu_{30} + \mu_{12})^2 + (\mu_{21} + \mu_{03})^2 \\
I_5 &= (\mu_{30} - 3\mu_{12})(\mu_{30} + \mu_{12})[(\mu_{30} + \mu_{12})^2 \\
& \ \ \ \ - 3(\mu_{21} + \mu_{03})^2] + (3\mu_{21} - \mu_{03})(\mu_{21} + \mu_{03}) \\
& \ \ \ \ * [3(\mu_{30} + \mu_{12})^2 - (\mu_{21} + \mu_{03})^2] \\
I_6 &= (\mu_{20} - \mu_{02})[(\mu_{30} + \mu_{12})^2 - (\mu_{21} + \mu_{03})^2] \\
& \ \ \ \ + 4\mu_{11}(\mu_{30} + \mu_{12})(\mu_{21} + \mu_{03}) \\
I_7 &= (3\mu_{21} - \mu_{03})(\mu_{30} + \mu_{12})[(\mu_{30} + \mu_{12})^2 \\
& \ \ \ \ - 3(\mu_{21} + \mu_{03})^2] -(\mu_{30} - 3 \mu_{12})(\mu_{21} + \mu_{03}) \\
& \ \ \ \ *[3(\mu_{30} + \mu_{12})^2 - (\mu_{21} + \mu_{03})^2]\end{aligned}
\end{equation}

\begin{equation}
\begin{aligned}
\label{eq:Hu-generate}
I_1 &\Leftrightarrow f(1,1) \\ 
I_2 &\Leftrightarrow (f(1,2))^2 - 2(g(1,2))^2 \\
I_3 &\Leftrightarrow (f(1,2))^3 - 3(g(1,2))^2f(1,2) \\
I_4 &\Leftrightarrow f(1,2)f(1,1)f(2,2) \\
I_5 &\Leftrightarrow f(2,2)f(3,3)f(4,4)[f(2,1)f(3,1)f(4,1) \\
& \ \ \ \ -f(2,1)g(3, 1)g(4,1)-g(2,1)g(3,1)f(4,1) \\
& \ \ \ \ -g(2,1)f(3,1)g(4,1)] \\
I_6 &\Leftrightarrow f(2,2)f(3,3)[f(1,2)f(1,3) - g(1,2)g(1,3)] \\
I_7 &\Leftrightarrow f(2,2)f(3,3)f(4,4)[g(2,1)f(3,1)f(4,1) \\
& \ \ \ \ -g(2,1)g(3,1)g(4,1)+f(2,1)g(3,1)f(4,1) \\
& \ \ \ \ +f(2,1)f(3,1)g(4,1)]
\end{aligned}
\end{equation}

From Eq.~\eqref{eq:Hu}, we can get the highest moment order and polynomial degree in an invariant. Similarly, we can draw the same conclusion from Eq.~\eqref{eq:Hu-generate} by the observation of the points. In the form of generating functions, the number of points involved and the occurrence of a point are important, because the highest occurrence for a point is the moment order defined by that point, and the number of points involved in a given invariant definition is the polynomial degree.
For example, there are two points in the definition of $I4$, and they both appear three times, so $I4$ is in order three, degree two.

We now explain in details how the order and degree are calculated for $I4$.
By plugging in Eq.~\eqref{eq:dot-product} and Eq.~\eqref{eq:cross-product}, we get:

\begin{equation}
\begin{aligned}
f(1,2) &= x_{1}x_{2} + y_{1}y_{2} \\
f(1,1) &= x_{1}x_{1} + y_{1}y_{1} \\
f(2,2) &= x_{2}x_{2} + y_{2}y_{2} 
\end{aligned}
\end{equation}

By introducing Eq.~\eqref{eq:central-moment}, we have

\begin{equation}
\begin{aligned}
I_{4} = & \int\int\int\int
(x_{1}x_{2} + y_{1}y_{2})
(x_{1}x_{1} + y_{1}y_{1})
(x_{2}x_{2} + y_{2}y_{2}) \\
& l(x_1,y_1)l(x_2,y_2)dx_1dy_1  dx_2dy_2
\end{aligned}
\end{equation}

For the definition $I_{4}$ involves two points $(x_1,y_1)$ and $(x_2,y_2)$, and the highest-power in the integrand is ($x_1^3x_2^3$), which defines a product of two moments of order three.

In general, an invariant may be a mixtured polynomial with different orders of moments. For example $I_6$ is a mixtured polymonial of order two and order three moments.

An interesting phenomenon is that different generating functions may generate the same invariant. Like $I_7$ can be decomposed into four different monomials of generating functions, but they are all equivalent invariants. We will explain it in details in the following section.

\section{Primitive Invariants}
\label{sec:PI}

In Eq.~\eqref{eq:Hu-generate}, we have re-expressed Hu's seven invariants in the form of generating functions. It is obvious that Hu's seven invariants can be decomposed into simpler monomials, which are the simplest moment invariants, and are referred to as Primitive Invariants (PIs).
Specifically, if the integral core of a moment invariant is a the product of a series of generating functions, we name it a Primitive Invariant.

From the polynomial in Eq.~\eqref{eq:Hu-generate}, those invariants can be decomposed into 16 PIs.
Further analysis shows that certain PIs are identical when translated to geometric moments invariants.
For example $I_5$ from Eq.~\eqref{eq:Hu-generate} can be re-expressed as the addition of four PIs, as shown in Eq.~\eqref{eq:I5}.

\begin{equation}
\begin{aligned}
\label{eq:I5}
I_5 &= I_{51} + I_{52} + I_{53} + I_{54} \\
where & \\
I_{51} &= f(2,2)f(3,3)f(4,4)f(2,1)f(3,1)f(4,1) \\
I_{52} &= f(2,2)f(3,3)f(4,4)f(2,1)g(3,1)g(4,1) \\
I_{53} &= f(2,2)f(3,3)f(4,4)g(2,1)g(3,1)f(4,1) \\
I_{54} &= f(2,2)f(3,3)f(4,4)g(2,1)f(3,1)g(4,1) \\
\end{aligned}
\end{equation}

It is easy to check that there exist the following linear relationships among these four PIs, shown in Eq.~\eqref{eq:I5_equation}.

\begin{equation}
\begin{aligned}
\label{eq:I5_equation}
I_{51} &= 3I_{52}\\ 
I_{52} &= I_{53} = I_{54}
\end{aligned}
\end{equation}

That is, there are essentially only two primitive invariants $I_{51}$ and $I_{52}$ in the decomposition of $I_5$. Similarly, $I_7$ can also be decomposed into four primitive invariants,
all of which are identical.

Hu \cite{Hu1962} pointed out that $I_7$, the last one from his seven invariants, is different from the first six invariants. $I_1$ through $I_6$ do not change their signs under reflection transformation, while $I_7$ does, which is called skew invariant.
Because of this, we specifically exclude $I_7$ from our proposed group of PIs.

From above, after simplification, there are 9 primitive invariants decomposed from Hu's first six invariants Eq.~\eqref{eq:Hu-generate}, shown in Eq.~\eqref{eq:PIs-decompos}.

\begin{equation}
\begin{aligned}
\label{eq:PIs-decompos}
I_{P1} &= f(1,1) \\ 
I_{P2} &= (f(1,2))^2 \\
I_{P3} &= (g(1,2))^2 \\
I_{P4} &= f(2,2)f(3,3)f(1,2)f(1,3)\\
I_{P5} &= \frac{1}{2}(g(1,2))^2f(1,2) \\
I_{P6} &= (f(1,2))^3 \\
I_{P7} &=  f(2,2)f(3,3)g(1,2)g(1,3) \\
I_{P8} &=  f(2,2)f(3,3)f(4,4)f(2,1)f(3,1)f(4,1) \\
I_{P9} &= f(2,2)f(3,3)f(4,4)f(2,1)g(3,1)g(4,1) \\
\end{aligned}
\end{equation}

Their corresponding invariants shown in Eq.~\eqref{eq:prime}.

\begin{equation}
\begin{aligned}
\label{eq:prime}
I_{P1} &= \mu_{20} + \mu_{02} \\
I_{P2} &= (\mu_{20})^2 + (\mu_{02})^2 + 2 (\mu_{11})^2 \\
I_{P3} &= \mu_{20}\mu_{02} - (\mu_{11})^2 \\
I_{P4} &= \mu_{20}(\mu_{30} + \mu_{12})^2 + 2\mu_{11}(\mu_{30} + \mu_{12})(\mu_{21} + \mu_{03}) \\
& \ \ \ \ + \mu_{02}(\mu_{21} + \mu_{03})^2 \\
I_{P5} &= \mu_{21}(\mu_{03} - \mu_{21}) + \mu_{12}(\mu_{30} - \mu_{12}) \\
I_{P6} &= (\mu_{30})^2 + 3(\mu_{21})^2 + 3(\mu_{12})^2 + (\mu_{03})^2 \\
I_{P7} &= \mu_{20}(\mu_{03} + \mu_{21})^2 - 2 \mu_{11}(\mu_{30} + \mu_{12})(\mu_{21} + \mu_{03}) \\
& \ \ \ \ + \mu_{02}(\mu_{30} + \mu_{12})^2 \\
I_{P8} &= \mu_{30}(\mu_{30} + \mu_{12})^3 + 3 (\mu_{30} + \mu_{12})(\mu_{21} + \mu_{03}) \\
& \ \ \ \ (\mu_{03} \mu_{12} + 2 \mu_{12} \mu_{21} + \mu_{21} \mu_{30}) + \mu_{03}(\mu_{21} + \mu_{03})^3 \\
I_{P9} &= \mu_{21}(\mu_{21} + \mu_{03})^3 - (\mu_{30} + \mu_{12})(\mu_{21} + \mu_{03}) \\
& \ \ \ \ (\mu_{03} \mu_{12} - 2 \mu_{03} \mu_{30} + 4 \mu_{12} \mu_{21} + \mu_{21} \mu_{30}) \\
& \ \ \ \ +\mu_{12}(\mu_{30} + \mu_{12})^3 \\
\end{aligned}
\end{equation}

Eq.~\eqref{eq:PIs-decompos} and Eq.~\eqref{eq:prime} are equivalent, with the former being able to be translated to the latter by using Eq.~\eqref{eq:central-moment}, Eq.~\eqref{eq:dot-product} and Eq.~\eqref{eq:cross-product}. This procedure is detailed in Eq.~\eqref{eq:Hu1_translation} of section \ref{sec:Translation}. 

Now Hu's seven invariants excluding the last one can be restructured using our 9 primitive invariants, shown in Eq.~\eqref{eq:Hu-prime1}.

\begin{equation}
\begin{aligned}
\label{eq:Hu-prime1}
I_1 &= I_{P1} \\
I_2 &= I_{P2} - 2I_{P3} \\
I_3 &= -6I_{P5} + I_{P6} \\
I_4 &= 2I_{P5} + I_{P6} \\
I_5 &= I_{P8} - 3I_{P9} \\
I_6 &= I_{P4} - I_{P7} \\
\end{aligned}
\end{equation}

It is interesting that Hu's original seven invariants can be reconstructed in a much simpler form using our primitive invariants. And nine PIs in total can be constructed, which, as a simpler set of invariants, will lead to less computational cost and does not include the skew invariant.
PIs provide a new perspective on the structure of invariants.

Note that the independence of the invariants are very important, the detailed explanations and proofs will be demonstrated in section \ref{sec:independent}. Here it is noticed that invariants in Eq.~\eqref{eq:Hu-prime1} are functional independent, and invariants in Eq.~\eqref{eq:prime} are also functional independent if eliminating $I_{P3}$, $I_{P7}$ and $I_{P9}$ from the group. Therefore, both groups of invariants can be used for simplified calculations.

%% file: Extension.tex

\section{Extension of Invariants}
In this section,two extensions are provided to show the generality of the generating functions in defining geometric moment invariants.

\subsection{3D Invariants}

First, the definition of generating functions in 2D space can be extended to 3D space,
shown in Eq.~\eqref{eq:dot-product3D} and Eq.~\eqref{eq:cross-product3D}.
The geometric meanings can be observed by definition of two functions, $f$ function represents the dot-product of two vectors, whereas $g$ function represents a third-order determinant, which is also the volume of a tetrahedron ignoring coefficients.

\begin{equation}
\label{eq:dot-product3D}
	f(i,j) = (x_{i},y_{i},z_{i})\cdot(x_{j},y_{j},z_{j}) = x_{i}x_{j} + y_{i}y_{j} + z_{i}z_{j}
\end{equation}

\begin{equation}
\begin{aligned}
\label{eq:cross-product3D}
	g(i,j,k) &=
	\begin{vmatrix}
x_{i} \ y_{i} \ z{i} \\
x_{j} \ y_{j} \ z{j} \\
x_{j} \ y_{j} \ z{k}
\end{vmatrix} \\
	&= x_{i}y_{j}z_{k} + x_{j}y_{k}z_{i} + x_{k}y_{i}z_{j} \\
	& \ \ - x_{i}y_{k}z_{j} - x_{j}y_{i}z_{k} - x_{k}y_{j}z_{i}
\end{aligned}
\end{equation}

Invariants are now constructed by careful selection of different combinations of generating functions and multiple integrals.
We demonstrate this straightforward procedure with the following example.

Three 3D invariants under rotation transformation in \cite{Sadjadi1980} are shown in Eq.~\eqref{eq:3D_3moments}.

\begin{equation}
\begin{aligned}
\label{eq:3D_3moments}
J_{1} &= \mu_{200} + \mu_{020} + \mu_{002} \\
J_{2}&= \mu_{200}\mu_{020}\mu_{002} + 2\mu_{110}\mu_{101}\mu_{011} - \mu_{011}^2 \\
& \ \ \ \ \ \mu_{200} - \mu_{110}^2\mu_{002} - \mu_{101}^2\mu_{020} \\
J_{3} &= \mu_{020}\mu_{002} - \mu_{011}^2 + \mu_{200}\mu_{002} - \mu_{101}^2  \\
& \ \ \ \ + \mu_{200}\mu_{020} - \mu_{110}^2 \\
\end{aligned}
\end{equation}

The definition of above invariants with our 3D generating functions are shown in Eq.~\eqref{eq:3D_3}.

\begin{equation}
\begin{aligned}
\label{eq:3D_3}
J_{1} &\Leftrightarrow f(1,1) \\
J_{2}&\Leftrightarrow g(1,2,3)^{2} \\
J_{3} &\Leftrightarrow f(1,1)f(2,2) - f(1,2)^2
\end{aligned}
\end{equation}

By combining Eq.~\eqref{eq:3D}, Eq.~\eqref{eq:dot-product3D} and Eq.~\eqref{eq:cross-product3D} with Eq.~\eqref{eq:3D_3}, it is trivial to prove that Eq.~\eqref{eq:3D_3moments} and Eq.~\eqref{eq:3D_3} are equivalent. This step is very similar to what was used in Eq.~\eqref{eq:Hu1_translation} of section \ref{sec:Translation} for the 2D space. 

Note that in Eq.~\eqref{eq:3D_3}, we have omitted the constant coefficients to keep the equations in a simple form. And for $J_3$, it is composed of two three-dimensional primitive invariants $f(1,1)f(2,2)$ and $f(1,2)^2$.

Therefore, we have demonstrated that the proposed method for generating invariants can be extended to 3D space. And by re-expressing the invariants using our generating functions, the results are equivalent to those in the form of geometric moments found in prior art, but in a much simpler and clearer format.

In theory, our method can be extended to higher dimensional space, which is a general way for constructing moment invariants. And it is also applicable to different manifolds such as 3D curve \cite{Li2006} and 3D surface \cite{Xu2006}.

\subsection{Affine Invariants}
\label{sec:affine}

Till now we have defined moment invariants all under similarity transformations, including rotation, translation and uniform scaling. 
And in fact, we do not directly discuss the uniform scaling, for we think it can be dealt with easy. As for translation, the use of central moment already get off its affect. In this section, we extend the transformation scope to affine transformation, which includes non-uniform scaling and is a general linear transformation.

For the generality of affine transformation, function $f$ no longer fits our need. Function $g$ does not hold its value after transformations either, however there exists a simple relationship for $g$ between those prior to the transformation and those afterwards, with their ratio being the determinant of the transformation. Therefore all affine moment invariants are defined only by function $g$. Some examples in 2D are listed below, but it can be extended to 3D space and different manifolds.

Flusser in \cite{Flusser2004} used graph method to represent and generate ten affine invariants in 2D. Here we re-express those ten invariants using our generating functions, as shown in Eq.~\eqref{eq:2D-affine}.
We only listed the numerators here,
and the denominators are in the form of $\mu_{00}^k$, where $\mu_{00}$ is defined in Eq.~\eqref{eq:central-moment}, and $k$ is related to the order and degree of invariants. In \cite{Flusser2004}, $k$ equals the sum of total number of nodes and edges in the corresponding graph.
In our method, $k$ equals the number of points plus the number of $g$ functions involved. For example, for $I_{A1}$, there are two points and two $g$ functions(square), its denominator power $k=4$.

In section \ref{sec:Translation} we have discussed the notion of order and degree for a given invariant. The order defines the highest occurrence of a certain point among all points, and the degree is the total number of points in a given invariant. These result also corresponds to Flusser's graph method, in which the notion of degree refers to the number of nodes in a graph, and the order refers to the number of edges in the graph.

\begin{equation}
\begin{aligned}
\label{eq:2D-affine}
I_{A1} &\Leftrightarrow (g(1,2))^2 \\
I_{A2} &\Leftrightarrow (g(1,2))^2(g(3,4))^2g(1,3)g(2,4) \\
I_{A3} &\Leftrightarrow g(1,2)g(1,3)(g(2,3))^2 \\
I_{A4} &\Leftrightarrow g(1,2)g(1,3)g(2,4)g(3,4)g(1,5)g(4,5) \\
I_{A5} &\Leftrightarrow (g(1,2))^4 \\
I_{A6} &\Leftrightarrow (g(1,2))^2(g(1,3))^2(g(2,3))^2 \\
I_{A7} &\Leftrightarrow (g(1,2))^2(g(1,3))^2 \\
I_{A8} &\Leftrightarrow (g(1,2))^2(g(2,3))^2(g(3,4))^2 \\
I_{A9} &\Leftrightarrow (g(1,2))^2g(2,3)(g(3,4))^2g(4,5)g(2,5)g(1,5) \\
I_{A10}&\Leftrightarrow (g(1,2))^4(g(3,4))^4g(1,3)g(2,4)
\end{aligned}
\end{equation}

We present a complete set of 19 affine invariants for practical use within order five. Eq.~\eqref{eq:2D-affine} includes the first ten invariants discussed by Flusser \cite{Flusser2004}. The remaining nine are shown in Eq.~\eqref{eq:2D-affine2}.
Their independence will be discussed in the next section.

\begin{equation}
\begin{aligned}
\label{eq:2D-affine2}
I_{A11} &\Leftrightarrow (g(1,2))^3(g(2,4))^2(g(3,4))^3 \\
I_{A12} &\Leftrightarrow g(1,2)(g(1,3))^2(g(2,3))^3 \\
I_{A13} &\Leftrightarrow g(1,2)(g(1,3))^3g(1,4)g(2,4)(g(3,4))^2 \\
I_{A14} &\Leftrightarrow g(1,3)g(1,4)(g(2,4))^2(g(3,4))^2 \\
I_{A15} &\Leftrightarrow (g(1,2))^3(g(1,3))^2(g(2,4))^2g(3,4) \\
I_{A16} &\Leftrightarrow g(1,2)g(1,3)(g(1,4))^2g(2,4)g(3,4) \\
I_{A17} &\Leftrightarrow g(1,2)g(1,3)g(1,4)(g(2,3))^2g(2,4) \\
I_{A18} &\Leftrightarrow (g(1,3))^2(g(2,4))^3g(3,4) \\
I_{A19} &\Leftrightarrow g(1,2)g(1,3)(g(2,3))^4
\end{aligned}
\end{equation}

The definition of our 19 affine invariants in the form of geometric moments are listed in the Appendix.

%% file: independent.tex

\section{Independent Invariants}
\label{sec:independent}

In section \ref{sec:PI} we have proved Hu's invariants can be decomposed into 9 primitive invariants. These invariants, as basic units in a much simpler form, providing a new perspective on the structure of invariants. The set of primitive invariants is promising to be used as a group of shape descriptors, only if they are functional independent of each other.
Functional independence is a stronger requirement than linear independence, because it might also include non-linear or transcendental functions. It is easy to prove linear independence of a group of vectors by checking the rank of its coefficients matrix. However it is usually difficult to prove functional independence, since there are still variables in its coefficients matrix. Hu's seven invariants are linear independent, but not functional independent. Flusser \cite{Flusser2000} has proved $I3$, $I5$ and $I7$ in Eq.~\eqref{eq:Hu-generate} are satisfying a nonlinear equation, they are not functional independent.

Those 9 primitive invariants may not necessarily be linear independent nor functional independent. From the definitions in Eq.~\eqref{eq:dot-product} and Eq.~\eqref{eq:cross-product}, the invariant core is the product of a series generating functions. Because only two generating functions are available to construct all moment invariants, the generated moment invariants are usually not functional independent of each other, thus they cannot be used as a group of shape descriptors.

The concept of independence of invariants is important, since a complete set of independent invariants would be able to describe the characteristics of a shape without information redundancy.

\subsection{Functional Independent Condition}
\label{sec:Background}

The notion of independent invariants was first introduced in \cite{Flusser2000} by Flusser to discuss invariants under rotation transformations. In his work, he reviewed invariants that had been proposed in the literature, and provided a way to select a group of invariants to form a complete set of independent invariants. However, the selected set is not a good shape descriptor, since it mixed true invariants (those do not change their values under reflection) with skew invariants (those change their values under reflection). As we explained in section \ref{sec:PI}, if placing those two types of invariants together as shape descriptors, it could lead to a large distance in characteristic space. This paper will propose a set of independent invariants all from true invariants group. 

In \cite{Brown1935}, Brown et al. proposed a technique to verify the functional independence for a group of functions. Based on that, we can establish the similar result for a group of invariants.

Suppose there are $m$ functions with $n$ parameters, $f_j(x_1,x_2,\cdots, x_n), j = 1,2,\cdots,m$. $J$ is the $m \times n$ matrix derived from  $f_j$'s first order partial derivatives, then we have followings.

\textbf{Theorem 1} (Functional Independence \cite{Brown1935}):
\label{FuncIndep}

For a group of functions $f_j(x_1,x_2,\cdots, x_n), j = 1,2,\cdots,m$, the necessary and sufficient condition for $f_j$ to be independent is that the rank of $J$ is $m$.

\textbf{Corollary 1}:
When $m > n$, the above group is functional dependent.

And since all the invariants are polynomial functions of moments, we have:

\textbf{Corollary 2}:
Given a certain moment order, the maximum number of independent invariants is determined by the total number of different moments.

For example, to define invariants under similarity transformations, including rotation, translation, and uniform scaling. We will show that there are maximum seven independent invariants within order three. This is because there are originally ten moments within order three, as listed in Eq.~\eqref{eq:moments_order_three}, 
\begin{equation}
\begin{aligned}
\label{eq:moments_order_three}
order \ 0: \ &m_{00} \\
order \ 1: \ &m_{10}, m_{01} \\
order \ 2: \ &m_{20}, m_{11}, m_{02} \\
order \ 3: \ &m_{30}, m_{21}, m_{12}, m_{03} \\
\end{aligned}
\end{equation}
where $m_{00}$ can be treated as a constant under similarity transformations,
$m_{10}$ and $m_{01}$ are constant zeros in the context of central moments.

Because moments at different orders are always independent of each other, we have seven different moments within order three, and with them we can get at most seven independent invariants.

This conclusion can be extended to higher orders or under different groups of transformations.

\subsection{Shape descriptors - a complete, simplest and independent set of invariants}
\label{sec:shape_descriptor}

In section \ref{sec:affine}, we have found 19 affine invariants. Here we prove these invariants form a complete, simplest and functional independent set of invariants.

\textbf{A complete set}:

Within order five, there are in total 21 moments. Under affine transformations, there are maximum 19 different moments. For $m_{10}$ and $m_{01}$ are constant zeros in the context of central moments as before, while $m_{00}$ becomes a variable, which counts for one of the independent moments. Note this is unlike the similarity transformations case where $m_{00}$ is treated as a constant. Therefore, 19 different moments indicate that there are maximum 19 independent invariants under the affine transformations within order five.

\textbf{A functional independent set}:

Based on Theorem 1 in Section \ref{FuncIndep}, we have verified that the proposed 19 affine invariants are functional independent from each other. Therefore they form a complete and independent set, and can be used as a group of shape descriptors.

The entire process of constructing invariants is as follows.

\begin{enumerate}
	\item Given the maximum number of involved points $N_{pnt}$, and the highest occurrence of one point $N_{cnt}$. Those would be the degree and the order in generated invariants, respectively;
	\item Enumerate all possible products of generating functions with $N_{pnt}$ and $N_{cnt}$;
	\item Translate each generating function to an invariant candidate;
	\item Go through each candidate and eliminate zero(s) and duplicates;
	\item Validate the remaining invariants with a maximum number of independent invariants.
\end{enumerate}

\textbf{The simplest set}: \\
We have shown that a complete set of invariants can be found, and those moment invariants are acquired by going through all possible combinations of generating functions within the given order.
However, in theory there could still be an infinite number of invariants even when the order of moments is fixed, because there will always be more invariants when the degree of invariants is increased.
Therefore, even if we know the maximum number of independent invariants within a certain order, the proposed set of invariants may not be the simplest set. Those invariants in higher degrees may also form an independent set, with or without invariants in lower degrees.    

In order to find a complete and simplest set of independent invariants, we propose to use the moments as lower order as possible. With the help of Maple software, we have tested that the set of affine invariants combining Eq.(26) and (27) is independent, with order five and degree five.

From our discussion above, we have the following propositions:

\textbf{Propositions 1}: For any given order of geometric moments, it is always possible to find a group of independent invariants by increasing the degree of 
them as polynomials of moments.

\textbf{Propositions 2}: It is always possible to find a group of independent invariants by increasing the order of geometric moments.

With our proposed method of construction of invariants, in theory there could be infinite invariants in higher order and/or degree. In practical scenarios, the lower order and/or degree of invariants are of importance, due to their noise robustness, less computational cost and improved reliability.

In summary, we start from lower order moments and seek for invariants by increasing the degree.
If the degree has been increased considerably but not enough candidates can be found, we gradually increase the order in order to include more moments in the construction of invariants.
When the maximum number of possible invariants is reached within certain order, we validate if the formed set of independent invariants is a complete and simplest set.
This procedure terminates when the proposed set has enough invariants to construct shape descriptors.

%% file: Conclusion.tex
\section{Conclusions}

In this paper we present a discovery that all geometric moment invariants (GMIs) can be defined by two fundamental generating functions.
Just like DNA encodes proteins for all living organisms, the two functions and their products form a minimalistic set of building blocks in the construction of geometric invariants, which are used to describe the characteristics of shapes.
A group of independent GMIs for affine transformations in 2D is proposed, some of which have not been mentioned in the literature.
The discovery of the two generating functions provides a new perspective to better understand shapes in 2D and 3D spaces, and can be extended in higher dimensional spaces and different manifolds.

%% file: Appendix.tex
\label{sec:Appendix}

In section \ref{sec:affine} we have shown our 19 affine invariants in the form of generating functions Eq.~\eqref{eq:2D-affine} and Eq.~\eqref{eq:2D-affine2}, here we show the definition of the invariants in the form of geometric moments. Those two forms are equivalent, and by re-expressing the invariants using our generating functions, the results are in a much simpler and clearer format. 

With the definition of geometric moments, we will show the complete form for our proposed affine invariants. In section \ref{sec:affine} we only listed the numerators, and the denominators are in the form of $\mu_{00}^k$ shown below, where $\mu_{00}$ is defined in Eq.~\eqref{eq:central-moment}, and $k$ is the sum of the number of points and the number of $g$ functions involved.

\clearpage

\begin{equation*}
\begin{aligned}
I_{A1} &= (\mu_{20}\mu_{02} - \mu_{11}^2)/\mu_{00}^4
\\
I_{A2} &= (-\mu_{03}^2\mu_{30}^2 + 6\mu_{03}\mu_{12}\mu_{21}\mu_{30} - 4\mu_{03}\mu_{21}^3 - 4\mu_{12}^3\mu_{30} + 3\mu_{12}^2\mu_{21}^2)/\mu_{00}^{10}
\\
I_{A3} &= (\mu_{02}\mu_{12}\mu_{30} - \mu_{02}\mu_{21}^2 - \mu_{03}\mu_{11}\mu_{30} + \mu_{03}\mu_{20}\mu_{21} + \mu_{11}\mu_{12}\mu_{21} - \mu_{12}^2\mu_{20})/\mu_{00}^{7}
\\
I_{A4} &= (\mu_{02}^3\mu_{30}^2 - 6\mu_{02}^2\mu_{11}\mu_{21}\mu_{30} + 3\mu_{02}^2\mu_{20}\mu_{21}^2 + 6\mu_{02}\mu_{11}^2\mu_{12}\mu_{30} + 6\mu_{02}\mu_{11}^2\mu_{21}^2 - 12\mu_{02}\mu_{11}\mu_{12}\mu_{20}\mu_{21}  \\
& \ \ \ + 3\mu_{02}\mu_{12}^2\mu_{20}^2 + \mu_{03}^2\mu_{20}^3 - 2\mu_{03}\mu_{11}^3\mu_{30} + 6\mu_{03}\mu_{11}^2\mu_{20}\mu_{21} - 6\mu_{03}\mu_{11}\mu_{12}\mu_{20}^2 - 6\mu_{11}^3\mu_{12}\mu_{21} + 6\mu_{11}^2\mu_{12}^2\mu_{20})/\mu_{00}^{11}
\\
I_{A5} &= (\mu_{04}\mu_{40} - 4\mu_{13}\mu_{31} + 3\mu_{22}^2)/\mu_{00}^{6}
\\
I_{A6} &= (\mu_{04}\mu_{22}\mu_{40} - \mu_{04}\mu_{31}^2 - \mu_{13}^2\mu_{40} + 2\mu_{13}\mu_{22}\mu_{31} - \mu_{22}^3)/\mu_{00}^{9}
\\
I_{A7} &= (\mu_{02}^2\mu_{40} - 4\mu_{02}\mu_{11}\mu_{31} + 2\mu_{02}\mu_{20}\mu_{22} + \mu_{04}\mu_{20}^2 + 4\mu_{11}^2\mu_{22} - 4\mu_{11}\mu_{13}\mu_{20})/\mu_{00}^{7}
\\
I_{A8} &= (\mu_{02}^2\mu_{22}\mu_{40} - \mu_{02}^2\mu_{31}^2 + \mu_{02}\mu_{04}\mu_{20}\mu_{40} - 2\mu_{02}\mu_{11}\mu_{13}\mu_{40} + 2\mu_{02}\mu_{11}\mu_{22}\mu_{31} - 2\mu_{02}\mu_{13}\mu_{20}\mu_{31} + \mu_{02}\mu_{20}\mu_{22}^2 \\ 
& \ \ \ - 2\mu_{04}\mu_{11}\mu_{20}\mu_{31} + \mu_{04}\mu_{20}^2\mu_{22} + 4\mu_{11}^2\mu_{13}\mu_{31} - 4\mu_{11}^2\mu_{22}^2 + 2\mu_{11}\mu_{13}\mu_{20}\mu_{22} - \mu_{13}^2\mu_{20}^2)/\mu_{00}^{10}
\\
I_{A9} &= (\mu_{03}^2\mu_{21}^2\mu_{40} - 2\mu_{03}^2\mu_{21}\mu_{30}\mu_{31} + \mu_{03}^2\mu_{22}\mu_{30}^2 - 2\mu_{03}\mu_{12}^2\mu_{21}\mu_{40} + 2\mu_{03}\mu_{12}^2\mu_{30}\mu_{31} - 2\mu_{03}\mu_{12}\mu_{13}\mu_{30}^2 \\
& \ \ \ + 2\mu_{03}\mu_{12}\mu_{21}^2\mu_{31} + 2\mu_{03}\mu_{13}\mu_{21}^2\mu_{30} - 2\mu_{03}\mu_{21}^3\mu_{22} + \mu_{04}\mu_{12}^2\mu_{30}^2 - 2\mu_{04}\mu_{12}\mu_{21}^2\mu_{30} + \mu_{04}\mu_{21}^4 \\
& \ \ \ +\mu_{12}^4\mu_{40} - 2\mu_{12}^3\mu_{21}\mu_{31} - 2\mu_{12}^3\mu_{22}\mu_{30}+ 2\mu_{12}^2\mu_{13}\mu_{21}\mu_{30} + 3\mu_{12}^2\mu_{21}^2\mu_{22} - 2\mu_{12}\mu_{13}\mu_{21}^3)/\mu_{00}^{13}
\\
I_{A10} &= (-\mu_{05}^2\mu_{50}^2 + 10\mu_{05}\mu_{14}\mu_{41}\mu_{50} - 4\mu_{05}\mu_{23}\mu_{32}\mu_{50} - 16\mu_{05}\mu_{23}\mu_{41}^2 + 12\mu_{05}\mu_{32}^2\mu_{41} - 16\mu_{14}^2\mu_{32}\mu_{50} \\
& \ \ \  - 9\mu_{14}^2\mu_{41}^2 + 12\mu_{14}\mu_{23}^2\mu_{50} + 76\mu_{14}\mu_{23}\mu_{32}\mu_{41} - 48\mu_{14}\mu_{32}^3 - 48\mu_{23}^3\mu_{41} + 32\mu_{23}^2\mu_{32}^2)/\mu_{00}^{14}
\\
I_{A11} &= (3\mu_{30}\mu_{23}^2\mu_{12} + 9\mu_{41}\mu_{23}\mu_{12}^2 + \mu_{23}\mu_{05}\mu_{30}^2 + 9\mu_{32}\mu_{14}\mu_{21}^2 + 3\mu_{21}\mu_{32}^2\mu_{03} + \mu_{50}\mu_{32}\mu_{03}^2 - \mu_{41}^2\mu_{03}^2 \\
& \ \ \  - 9\mu_{32}^2\mu_{12}^2 - 9\mu_{23}^2\mu_{21}^2 - \mu_{14}^2\mu_{30}^2 - \mu_{50}\mu_{03}\mu_{05}\mu_{30} + 3\mu_{50}\mu_{03}\mu_{14}\mu_{21} - 3\mu_{50}\mu_{03}\mu_{23}\mu_{12} + 3\mu_{41}\mu_{12}\mu_{05}\mu_{30} \\
& \ \ \  + 2\mu_{41}\mu_{03}\mu_{14}\mu_{30} + 3\mu_{41}\mu_{12}\mu_{32}\mu_{03} - 6\mu_{41}\mu_{03}\mu_{23}\mu_{21} - 9\mu_{41}\mu_{12}\mu_{14}\mu_{21} - 3\mu_{32}\mu_{21}\mu_{05}\mu_{30} \\
& \ \ \  + 9\mu_{32}\mu_{21}\mu_{23}\mu_{12} - \mu_{32}\mu_{03}\mu_{23}\mu_{30} - 6\mu_{32}\mu_{12}\mu_{14}\mu_{30} + 3\mu_{23}\mu_{30}\mu_{14}\mu_{21})/\mu_{00}^{12}
\\
I_{A12} &= (\mu_{50}\mu_{13}\mu_{03} - \mu_{50}\mu_{04}\mu_{12} - 3\mu_{41}\mu_{22}\mu_{03} + \mu_{41}\mu_{13}\mu_{12} + 2\mu_{41}\mu_{04}\mu_{21} + 3\mu_{32}\mu_{31}\mu_{03} - 5\mu_{32}\mu_{13}\mu_{21} \\
& \ \ \  + 3\mu_{32}\mu_{22}\mu_{12} - \mu_{32}\mu_{04}\mu_{30} - \mu_{23}\mu_{40}\mu_{03} - 5\mu_{23}\mu_{31}\mu_{12} + 3\mu_{23}\mu_{22}\mu_{21} + 3\mu_{23}\mu_{13}\mu_{30} + 2\mu_{14}\mu_{40}\mu_{12} \\
& \ \ \  + \mu_{14}\mu_{31}\mu_{21} - 3\mu_{14}\mu_{22}\mu_{30} - \mu_{05}\mu_{40}\mu_{21} + \mu_{05}\mu_{31}\mu_{30})/\mu_{00}^{9}
\\
I_{A13} &= (-\mu_{02}\mu_{05}\mu_{31}\mu_{50} + \mu_{02}\mu_{05}\mu_{40}\mu_{41} + \mu_{02}\mu_{14}\mu_{22}\mu_{50} + 3\mu_{02}\mu_{14}\mu_{31}\mu_{41} - 4\mu_{02}\mu_{14}\mu_{32}\mu_{40} - 4\mu_{02}\mu_{22}\mu_{23}\mu_{41} \\
& \ \ \  + 3\mu_{02}\mu_{22}\mu_{32}^2 + 3\mu_{02}\mu_{23}^2\mu_{40} - 2\mu_{02}\mu_{23}\mu_{31}\mu_{32} + \mu_{04}\mu_{14}\mu_{20}\mu_{50} - 4\mu_{04}\mu_{20}\mu_{23}\mu_{41} + 3\mu_{04}\mu_{20}\mu_{32}^2 \\
& \ \ \  + 2\mu_{05}\mu_{11}\mu_{22}\mu_{50} - 2\mu_{05}\mu_{11}\mu_{31}\mu_{41} - \mu_{05}\mu_{13}\mu_{20}\mu_{50} + \mu_{05}\mu_{20}\mu_{22}\mu_{41} - 2\mu_{11}\mu_{13}\mu_{14}\mu_{50} + 8\mu_{11}\mu_{13}\mu_{23}\mu_{41} \\
& \ \ \   - 6\mu_{11}\mu_{13}\mu_{32}^2 - 6\mu_{11}\mu_{14}\mu_{22}\mu_{41} + 8\mu_{11}\mu_{14}\mu_{31}\mu_{32} + 4\mu_{11}\mu_{22}\mu_{23}\mu_{32} - 6\mu_{11}\mu_{23}^2\mu_{31} + 3\mu_{13}\mu_{14}\mu_{20}\mu_{41}  \\
& \ \ \  - 2\mu_{13}\mu_{20}\mu_{23}\mu_{32} - 4\mu_{14}\mu_{20}\mu_{22}\mu_{32} + 3\mu_{20}\mu_{22}\mu_{23}^2)/\mu_{00}^{12}
\\
I_{A14} &= (\mu_{02}^2\mu_{12}\mu_{50} - 2\mu_{02}^2\mu_{21}\mu_{41} + \mu_{02}^2\mu_{30}\mu_{32} - \mu_{02}\mu_{03}\mu_{11}\mu_{50} + \mu_{02}\mu_{03}\mu_{20}\mu_{41} - \mu_{02}\mu_{11}\mu_{12}\mu_{41}  \\
& \ \ \  + 5\mu_{02}\mu_{11}\mu_{21}\mu_{32} - 3\mu_{02}\mu_{11}\mu_{23}\mu_{30} - \mu_{02}\mu_{12}\mu_{20}\mu_{32} + \mu_{02}\mu_{14}\mu_{20}\mu_{30} - \mu_{02}\mu_{20}\mu_{21}\mu_{23} + 2\mu_{03}\mu_{11}^2\mu_{41} \\ 
& \ \ \  - 3\mu_{03}\mu_{11}\mu_{20}\mu_{32} + \mu_{03}\mu_{20}^2\mu_{23} - \mu_{05}\mu_{11}\mu_{20}\mu_{30} + \mu_{05}\mu_{20}^2\mu_{21} - 2\mu_{11}^2\mu_{12}\mu_{32} + 2\mu_{11}^2\mu_{14}\mu_{30} - 2\mu_{11}^2\mu_{21}\mu_{23} \\
& \ \ \  + 5\mu_{11}\mu_{12}\mu_{20}\mu_{23} - \mu_{11}\mu_{14}\mu_{20}\mu_{21} - 2\mu_{12}\mu_{14}\mu_{20}^2)/\mu_{00}^{10}
\\
I_{A15} &= (- \mu_{03}\mu_{05}\mu_{30}\mu_{50} + 2\mu_{03}\mu_{14}\mu_{21}\mu_{50} + 3\mu_{03}\mu_{14}\mu_{30}\mu_{41} - 8\mu_{03}\mu_{21}\mu_{23}\mu_{41} + 6\mu_{03}\mu_{21}\mu_{32}^2 - 2\mu_{03}\mu_{23}\mu_{30}\mu_{32} \\
& \ \ \  + \mu_{05}\mu_{12}\mu_{21}\mu_{50} + 2\mu_{05}\mu_{12}\mu_{30}\mu_{41} - 2\mu_{05}\mu_{21}^2\mu_{41} - 2\mu_{12}^2\mu_{14}\mu_{50} + 8\mu_{12}^2\mu_{23}\mu_{41} - 6\mu_{12}^2\mu_{32}^2 \\
& \ \ \  - 3\mu_{12}\mu_{14}\mu_{21}\mu_{41} - 8\mu_{12}\mu_{14}\mu_{30}\mu_{32} + 2\mu_{12}\mu_{21}\mu_{23}\mu_{32} + 6\mu_{12}\mu_{23}^2\mu_{30} + 8\mu_{14}\mu_{21}^2\mu_{32} - 6\mu_{21}^2\mu_{23}^2)/\mu_{00}^{12}
\\
I_{A16} &= (-2\mu_{40}\mu_{11}\mu_{02}\mu_{13} + \mu_{40}\mu_{11}^2\mu_{04} + \mu_{40}\mu_{02}^2\mu_{22} - 2\mu_{31}\mu_{20}\mu_{11}\mu_{04} + 2\mu_{31}\mu_{11}\mu_{02}\mu_{22} - \mu_{02}^2\mu_{31}^2 \\ 
& \ \ \ + 2\mu_{31}\mu_{20}\mu_{02}\mu_{13} + 2\mu_{22}\mu_{20}\mu_{11}\mu_{13} + \mu_{22}\mu_{20}^2\mu_{04} - \mu_{11}^2\mu_{22}^2 - 2\mu_{20}\mu_{02}\mu_{22}^2 - \mu_{20}^2\mu_{13}^2)/\mu_{00}^{10}
\\
I_{A17} &= (\mu_{20}\mu_{30}\mu_{13}\mu_{03} + \mu_{20}\mu_{04}\mu_{21}^2 - \mu_{20}\mu_{21}\mu_{22}\mu_{03} - \mu_{20}\mu_{21}\mu_{13}\mu_{12} + \mu_{20}\mu_{22}\mu_{12}^2 - \mu_{20}\mu_{30}\mu_{04}\mu_{12} \\
& \ \ \  + 2\mu_{11}\mu_{21}\mu_{22}\mu_{12} + 2\mu_{11}\mu_{21}\mu_{31}\mu_{03} + 2\mu_{11}\mu_{12}\mu_{13}\mu_{30} - 2\mu_{11}\mu_{30}\mu_{22}\mu_{03} - 2\mu_{11}\mu_{31}\mu_{12}^2 - 2\mu_{11}\mu_{13}\mu_{21}^2 \\
& \ \ \  + \mu_{02}\mu_{22}\mu_{21}^2 - \mu_{02}\mu_{21}\mu_{31}\mu_{12} + \mu_{02}\mu_{40}\mu_{12}^2 - \mu_{02}\mu_{03}\mu_{40}\mu_{21} + \mu_{02}\mu_{03}\mu_{31}\mu_{30} - \mu_{02}\mu_{30}\mu_{22}\mu_{12})/\mu_{00}^{10}
\\
I_{A18} &= (\mu_{02}\mu_{03}\mu_{40}\mu_{21} - 2\mu_{40}\mu_{12}\mu_{03}\mu_{11} + \mu_{40}\mu_{03}^2\mu_{20} - 3\mu_{02}\mu_{21}\mu_{31}\mu_{12} + 2\mu_{11}\mu_{21}\mu_{31}\mu_{03} - \mu_{02}\mu_{03}\mu_{31}\mu_{30} \\
& \ \ \  + 6\mu_{11}\mu_{31}\mu_{12}^2 - 4\mu_{31}\mu_{12}\mu_{03}\mu_{20} + 3\mu_{02}\mu_{30}\mu_{22}\mu_{12} - 12\mu_{11}\mu_{21}\mu_{22}\mu_{12} + 3\mu_{20}\mu_{22}\mu_{12}^2 + 3\mu_{02}\mu_{22}\mu_{21}^2 \\
& \ \ \  + 3\mu_{20}\mu_{21}\mu_{22}\mu_{03} - 4\mu_{13}\mu_{30}\mu_{21}\mu_{02} + 6\mu_{11}\mu_{13}\mu_{21}^2 - \mu_{20}\mu_{30}\mu_{13}\mu_{03} + 2\mu_{11}\mu_{12}\mu_{13}\mu_{30} - 3\mu_{20}\mu_{21}\mu_{13}\mu_{12} \\
& \ \ \  + \mu_{04}\mu_{30}^2\mu_{02} - 2\mu_{04}\mu_{21}\mu_{30}\mu_{11} + \mu_{20}\mu_{30}\mu_{04}\mu_{12})/\mu_{00}^{10}
\\
I_{A19} &= (\mu_{02}\mu_{14}\mu_{50} - 4\mu_{02}\mu_{23}\mu_{41} + 3\mu_{02}\mu_{32}^2 - \mu_{05}\mu_{11}\mu_{50} + \mu_{05}\mu_{20}\mu_{41} + 3\mu_{11}\mu_{14}\mu_{41} - 2\mu_{11}\mu_{23}\mu_{32} \\
& \ \ \  - 4\mu_{14}\mu_{20}\mu_{32} + 3\mu_{20}\mu_{23}^2)/\mu_{00}^{9}
\\
\end{aligned}
\end{equation*}

\clearpage